\documentclass[english]{llncs}
\usepackage[T1]{fontenc}
\usepackage[latin9]{inputenc}
\usepackage{fancyhdr}
\usepackage{units}
\usepackage{bm}
\usepackage{amsmath}
\usepackage{amssymb}
\usepackage{graphicx}

\makeatletter

\providecommand{\tabularnewline}{\\}

\usepackage{bm}

\makeatother

\usepackage{babel}
\begin{document}

\title{Estimating the intrinsic dimension in fMRI space via dataset fractal
analysis}

\subtitle{Counting the `cpu cores' of the human brain}

\titlerunning{Estimating the intrinsic dimension in fMRI space via dataset fractal
analysis}

\author{Harris V. Georgiou (MSc, PhD)}

\authorrunning{Harris V. Georgiou}

\institute{Dept. of Informatics \& Telecommunications,\\
National Kapodistrian Univ. of Athens (NKUA/UoA)\\
\bigskip{}
\email{Email: harris@xgeorgio.info --- URL: http://xgeorgio.info}}
\maketitle
\begin{abstract}
Functional Magnetic Resonance Imaging (fMRI) is a powerful non-invasive
tool for localizing and analyzing brain activity. This study focuses
on one very important aspect of the functional properties of human
brain, specifically the estimation of the level of \emph{parallelism}
when performing complex cognitive tasks. Using fMRI as the main modality,
the human brain activity is investigated through a purely data-driven
signal processing and dimensionality analysis approach. Specifically,
the fMRI signal is treated as a multi-dimensional data space and its
intrinsic `complexity' is studied via \emph{dataset fractal analysis}
and \emph{blind-source separation} (BSS) methods. One simulated and
two real fMRI datasets are used in combination with Independent Component
Analysis (ICA) and fractal analysis for estimating the intrinsic (true)
dimensionality, in order to provide data-driven experimental evidence
on the number of independent brain processes that run in parallel
when visual or visuo-motor tasks are performed. Although this number
is can not be defined as a strict threshold but rather as a continuous
range, when a specific activation level is defined, a corresponding
number of parallel processes or the casual equivalent of `cpu cores'
can be detected in normal human brain activity.
\end{abstract}

\section{Introduction \label{sec:intro} }

Human brain is the most advanced and efficient signal-processing machine
known today. It corresponds to only 2\% of total body weight in adults
(about 1.5 kg), yet it consumes 20\% of blood oxygen and 25\% of glucose,
with only 20W at power peak. It consists of roughly 100 billion neurons
with 1,000-10,000 synapse interconnections each, packed in 1130-1260
cm\textsuperscript{3} of volume, making it the most complex organ
in the human body \cite{brainbookC_1995,Cosgrove_brain_2007,lazarbook}.
Analyzing its structure and functionality, especially during the actual
process of some cognitive task or in relation to some mental impairment,
has been a scientific challenge for centuries. However, only recent
technological breakthroughs have enabled the study of the inner workings
of living brains. Even today, simulating the structure and only basic
neuron functionality of a full-scale human brain in a digital computer
is still an infeasible task.

Functional Magnetic Resonance Imaging (fMRI) \cite{lindquist_statistical_2008,lazarbook,smith_fmri_2004}
is a powerful non-invasive tool for localizing and analyzing brain
activity. Most commonly it is based on blood oxygenation level-dependent
(BOLD) contrast, which translates to detecting localized changes in
the hemodynamic flow of oxygenated blood in activated brain areas.
This is achieved by exploiting the different magnetic properties of
oxygen-saturated versus oxygen-desaturated hemoglobin.

In the human brain, tasks involving action, perception, cognition,
etc., are performed via the simultaneous activation of a number of
\textit{functional brain networks} (FBN), which are engaged in proper
interactions in order to effectively execute the task. Such networks
are usually related to low-level brain functions and they are defined
as a number of \textit{segregated} specialized small brain regions,
potentially distributed over the entire brain. In order to properly
detect these activations and identify the set regions that constitutes
a FBN related to a specific task, the 3-D space occupied by the brain
is partitioned into a grid of `cubes' or \emph{voxels}. Each voxel
constitutes the elementary spatial unit that acts as a signal generator,
recorded and registered as a low-resolution 1-D time series. Actual
fMRI voxel signals from brain scans can be considered as a mixture
of various components or \emph{sources} with different temporal and
spatial characteristics. These sources can be classified as of interest
and as artifacts \cite{calhoun_ica_2003}.

In order to understand the true functionality and full potential of
the human brain, data-intensive approaches are required for analyzing
the actual brain signal (e.g. fMRI, EEG) during specific cognitive
tasks. Current research involves multi-disciplinary endeavors, from
Biochemistry and Neurophysiology to Simulation and VLSI design, with
projects like the Human Brain Project (HBP) by EU \cite{HBP-EU-url}
and Brain Research through Advancing Innovative Neurotechnologies
(BRAIN) by USA \cite{BRAIN-USA-url}. There is also very active research
and development effort in the industry, where projects like the recently
announced `TrueNorth' chip by IBM implement a million-scale neural
network grid in special-purpose VLSI and extremely high power efficiency
\cite{Merolla_truenorth_2014,Sebastian_truenorth_2014}. However,
all these efforts are currently focused on the \emph{structural} properties
of the human brain, i.e., the neural networks topology and connectivity,
while the \emph{functional} and higher-level cognitive properties
are still very difficult to model. In practice, this means that the
hardware necessary to build and fully simulate (at the neuron cell
level) an actual artificial `brain' equivalent to a small animal's
now becomes available, but the problem of turning this construction
to a machine with actual cognitive and abstract functionality still
remains (for the most part) unsolved, with only application-specific
modules being developed successfully (e.g. artificial retina implants,
with some visual processing capabilities \cite{ArtifRetina-USA-url}).

This study focuses on one very important aspect of the functional
properties of human brain, specifically the estimation of the level
of \emph{parallelism} when performing complex cognitive tasks. In
some very abstract sense, this is not much different than trying to
recover the (minimum) number of actual `cpu cores' required to `run'
all the active cognitive tasks that are registered in the entire 3-D
brain volume while performing a typical fMRI experimental protocol
that includes visual-only or visuo-motor tasks.

Using fMRI as the main modality, the human brain activity is investigated
through a purely data-driven signal processing and dimensionality
analysis approach. Specifically, the fMRI signal is treated as a multi-dimensional
data space and its intrinsic complexity is studied via \emph{dataset
fractal analysis} and \emph{blind-source separation} methods. Section
\ref{sub:intro-fmri-data} provides an overview of the fMRI experiments
and the nature of sensory data; section \ref{sub:intro-brain-activity}
defines a proper mathematical formulation for the \emph{data unmixing}
task and its importance in understanding the true sources of brain
activity; section \ref{sub:decimation-dimension} provides hints to
proper data dimensionality reduction in fMRI; section \ref{sub:dataset-fractal-analysis}
briefly describes the basic methodology for dataset fractal analysis
and how it is applied for the estimation of the \emph{intrinsic dimensionality}
of the fMRI data space; section \ref{sub:ICA-descr} briefly describes
ICA as a typical approach for blind-source separation in signal processing;
sections \ref{sub:Simulated-fMRI-datasets}, \ref{sub:ds101-descr}
and \ref{sub:ds105-descr} describe the simulated and real fMRI datasets
used in this study; section \ref{sec:Experim-Results} includes the
experiments and results, using all the methods and datasets described
earlier; sections \ref{sec:Discussion} and \ref{sec:Conclusion}
conclude the study with discussion of the results and their practical
meaning.

\section{Problem Definition}

\subsection{The nature of fMRI data \label{sub:intro-fmri-data}}

In experimental fMRI procedures, there are two common activation schemes:
the \emph{block} paradigms and the \emph{event-related} paradigms
\cite{behroozi_fmri_2011}. In the block paradigm, the subject is
presented with a specific stimulus for a specific time frame, e.g.,
a set of images of different placement, colors, patterns or categories,
and the subject has to press a switch to signal positive or negative
feedback as a response. In the event-related paradigm, the subject
is exposed to a series of randomized short-time inputs, e.g., a noise
or a pain stimulus, with or without the need for specific response
from the subject. In both cases, the external input is considered
as a primary `source' and is temporally correlated with the brain
activity. Areas of high activation and correlation to the stimulation/response
pattern are considered as highly relevant to the specific functional
task (visual/motor centers, pain receptors, etc). The same procedure
can be followed when there is no specific external paradigm, constituting
the \emph{steady-state} functional analysis of brain activity. In
this setup, there is no correlation to previously known activation
pattern and hence the analysis is essentially a search for functionally
independent sources in the recorded fMRI signal.

The acquired fMRI signal is registered in both spatial (3-D) and temporal
(1-D) domain, resulting in a composite 4-D signal. Each spatial axis
is registered as a grid of spatial resolution 3-5 mm\textsuperscript{3},
resulting in a 3-D grid of voxels. Typically, a complete volume of
voxel data, e.g. 60x60x30 to 64x64x48, is recorded every 1-2 seconds
for a sequence of 100-150 time points \cite{lindquist_statistical_2008,lazarbook,smith_fmri_2004}.
This produces a total of roughly 108K-197K voxels for every time frame
or, equivalently, 11e6 to 30e6 data points organized as a two-dimensional
matrix, where each row corresponds to a complete brain `snapshot'.
In practice, the number of actual brain voxels is smaller, since non-brain
areas of the grid are masked out before any further processing; however,
the data volume still remains within the same order of magnitude.
Additionally, typical fMRI experimental protocols involve several
subjects, in order to exclude any subject-specific characteristics
that may affect the statistical properties of the fMRI data under
consideration. Clearly, this creates a high-volume data analysis process
that makes it a very complex and computationally demanding task.

In terms of signal processing, the hemodynamic response function (HRF)
\cite{lindquist_statistical_2008,lazarbook,smith_fmri_2004} of the
activated neurons, i.e., the changes in oxygen-rich blood flow in
the time domain, acts as a low-pass filter in the temporal domain,
which in turn modifies the true activation signal that it is registered
as fMRI data. In other words, the HRF of the activated neurons, i.e.,
the changes in oxygen-rich blood flow in the time domain, modifies
the true activation time-series signal that it is registered for each
brain voxel as fMRI data. Moreover, the HRF is known to be spatially-varying,
which means that there are slightly different hemodynamic responses
for different areas of the brain, as well as different HRFs between
different subjects. This means that traditional regression approaches
like General Linear Model (GLM) approximations \cite{lindquist_statistical_2008,smith_fmri_2004,lazarbook}
that require a pre-defined `design matrix' are clearly sub-optimal,
since it is typically constructed as permutations, transformations,
time-shifts and derivatives of one (assumed) `universal' HRF. There
are also additional features that makes this approximation even more
difficult in practice, such as the fact that the voxels' activations
are assumed to be statistically independent (while locally they are
not, due to the physical properties of the veins and hemodynamics),
as well the various artifacts that are introduced to the signal by
external factors (scanner drift, electronic noise, head movements,
respiration, cardiac pulsation, etc) \cite{lindquist_statistical_2008,lazarbook,smith_fmri_2004}.

\subsection{Understanding brain activity \label{sub:intro-brain-activity}}

In fMRI analysis, the sources of interest include task-related, transiently
task-related and function-related sources, meaning that in a task-specific
fMRI experiment most of the task-related activity is expected to be
spatially isolated and temporally synchronized with the corresponding
input/stimulation patterns. Therefore, these sources are expected
to appear as super-Gaussian in nature due to the spatial and temporal
localization of such task-related brain functionality. 

Special matrix factorization algorithms are required to reformulate
the fMRI data as a multiplication of two other matrices, where one
is for the time courses of the estimated signal `sources' and one
for the corresponding spatial maps of related brain activity. Formally
put, if $\bm{Y}\in\mathbb{R}^{t\times n}$ is the full fMRI data matrix
with $t$ rows as time points and $n$ brain voxels `unwrapped' into
a linear vector, then the fMRI data matrix can be factorized as $\bm{Y}=\bm{T}\bm{S}$,
$\bm{T}\in\mathbb{R}^{t\times p}$, $\bm{S}\in\mathbb{R}^{p\times n}$,
where the $p$ spatial maps are collected as rows in $\bm{S}$ and
each column of $\bm{T}$ contains the activation pattern along time
for the corresponding spatial map.

In GLM \cite{lindquist_statistical_2008,smith_fmri_2004,lazarbook},
$\bm{T}$ is the pre-defined `design matrix' that contains permutations,
transformations, time-shifts and derivatives of one (assumed) `universal'
HRF, while $\bm{S}$ is the matrix of the corresponding regressor
coefficients in each row. Although the GLM approach is sufficient
when only specific sensory-related signal sources (external stimuli)
are considered, in the general case it is not possible to define a
global design matrix for all signal sources and all (multiple) subjects.
Instead, Independent Component Analysis (ICA) is the most commonly
used method for this task, in the context of blind source separation
(BSS) \cite{hyvarbook,calhoun_stica_2001,jung_introduction_2001}
(see section \ref{sub:ICA-descr}).

While GLM and ICA are the dominating approaches for directed or blind
unmixing models, respectively, in fMRI analysis, the large volume
of voxel data and the inherent properties of the fMRI signals make
the unmixing task highly demanding in both memory and computational
resources. Moreover, proper identification of `universal' FBNs requires
multiple experiments with different subjects, which means working
with multiplied volume of sensory data or combining multiple unmixing
results over various runs \cite{Erhardt2011,chen_linear_2013}. In
either case, unmixing algorithms are required to be both fast and
accurate in identifying the signal `sources' of fMRI data and the
activated areas in the brain corresponding to the specific paradigm
source.

Based on these properties, it is clear that the analysis of fMRI data,
in the sense of its decomposition into distinct sources and the identification
of the ones related to a specific task or functional activity in the
brain, is a very difficult task. The lack of strict specifications
for a `universal' HRF and background artifacts, hence in turn for
an accurate pre-defined `design matrix' for a standard GLM model,
makes it a typical candidate for BSS approaches such as ICA, as well
as alternative approaches like Dictionary Learning (DL) and Compressive
Sensing (CS). Recently, there is an increased interest for alternatives
to ICA for data-driven fMRI unmixing. Notably good results have been
attained with Dictionary Learning (DL) - based fMRI analysis \cite{kangjoo_lee_data-driven_2011,abolghasemi_fast_2013,abrahamsen_sparse_2011}.
Also, an improved variation of K-SVD was proposed as the basis for
Dictionary Learning (DL), customized to fMRI analysis \cite{Eusipco2014}.

\section{Dimensionality analysis of the fMRI data space}

\subsection{Data decimation and intrinsic dimensionality \label{sub:decimation-dimension}}

One way to deal with the high complexity of the BSS task in fMRI data
is to reduce the number of voxels under consideration. Specifically,
adjacent neurons in the brain can be considered highly correlated
in terms of their responses to external stimuli, provided that the
blood vessel networks at very small scales actually introduce some
spatio-temporal correlation. Hence, their BOLD response and HRF can
be considered, at some degree, statistically dependent. If the spatial
resolution of the fMRI signal is high, adjacent voxels in the original
2-D or 3-D volume scan can be considered statistically redundant.
Therefore, some form of decimated voxels set can be used instead as
input for the unmixing task, without sacrificing the accuracy of identifying
the true inherent sources of the data.

In fMRI, using decimated versions of the original data in the spatial
and/or the temporal domain is not an uncommon practice. Indeed, some
works refer to sub-sampling the fMRI in a spatial (voxels) or temporal
(time points) sense, when the resolution is considered high enough.
However, this has been applied only as a one-time pre-processing step
in the preparation of data, i.e., before any real BSS analysis is
conducted (e.g. see in \cite{Lee_ddfmri_2010,kangjoo_lee_data-driven_2011}).
Furthermore, the decimation ratio used in each case is chosen in a
purely empirical way, since there are currently no analytical studies
with regard to the resulting quality of the decimated fMRI data.

Spatio-temporal correlations between voxels and statistical dependencies
are essentially the reason why the fMRI data space has an intrinsic
(true) dimensionality much smaller than the number of voxels, i.e.,
the data matrix $\bm{Y}\in\mathbb{R}^{t\times n}$ is of column rank
$c\ll n$. However, for proper unmixing of the fMRI data, the column
rank of matrix $\bm{Y}$ should be retained even when some decimation
process is employed. In other words, the selection of a smaller subset
of voxels (instead of all) should be conducted in a way that does
not destroy the information content of the full data, but instead
exploit the the fact that the number of voxels $n$ is very large
and their inherent statistical properties can be properly retained
with a much smaller subset.

In the cases when only a small set of the signal sources are considered,
i.e., the time series of some external stimuli (plus some transformations
of it), then regression methods like GLM can be easily formulated
with the proper `design matrix' to recover the related brain activity.
When the analysis is conducted in the BSS sense, i.e., all major signal
sources are to be recovered (including the stimuli time series), then
decomposition methods like ICA provide a well-formulated statistical
framework for this task, as long as the proper constraints are asserted
as valid (most importantly, the assertion of at most one Gaussian
signal source). However, when these statistical assertions are not
fully satisfied or when there is a large number of signal sources
that are `exponentially decaying' in terms of importance (contribution
to the mixed signal's variance, power spectrum and approximation error),
then the number of independent components that ICA or other similar
algorithms is limited only by some external pre-defined threshold.
In other words, the data matrix $\bm{Y}\in\mathbb{R}^{t\times n}$
can be factorized \emph{approximately} as $\bm{Y}\simeq\bm{T}\bm{S}$,
$\bm{T}\in\mathbb{R}^{t\times p}$, $\bm{S}\in\mathbb{R}^{p\times n}$,
with the reconstruction error becoming smaller as $p$ increases.
In theory, if the true sources of the mixed signal are perfectly separable
in the BSS sense, then ICA will stop after recovering exactly $p=c$
components, where $c\ll n$ is the column rank of the data matrix
$\bm{Y}$. This means that there are exactly $p$ components, i.e.
time courses and corresponding activation maps, that can fully reconstruct
the fMRI data for the entire brain activity. Hence, the definition
of the optimal value for $p$ by means of non-parametric (data-driven)
estimation procedure is of utmost importance in the BSS task for fMRI
unmixing.

\subsection{Dataset fractal analysis \label{sub:dataset-fractal-analysis}}

In recent years, dimensionality analysis in signal processing has
been extensively linked to fractal analysis and \emph{fractal dimension},
as a non-parametric method for the quantitative characterization of
the complexity or `randomness' of a signal \cite{fracbookg_1992,fracbookw_1994}.
When applied to 1-D signals, metrics like the \emph{Hurst exponent}
or \emph{Lyapunov exponent} have been used as statistical features
to describe various types of data series, from biomedical signals
(e.g. EEG, ECG, etc) to financial and climate time series. In 2-D
signals, these methods provide additional features for characterizing
the texture of images, e.g. when analyzing biomedical modalities (radiology,
ultrasound, MRI, etc) \cite{chen_fracanal_1989}. Fractal dimension
is closely linked to these fractal parameters and it provides a clear
distinction between the \emph{embedding} space, i.e., the full-rank
space in the algebraic sense, from the actual space spanned by the
registered sensory data. In the general case when fractal analysis
is applied to some multi-dimensional signal, the estimation of the
fractal dimension can be used as a realistic evaluation of the `complexity'
of the space spanned by the actual data points available and, hence,
a very useful hint regarding the inherent redundancy in a given dataset. 

In order to establish a preliminary estimation of the complexity and
intrinsic dimensionality of datasets, fractal analysis provides a
data-centric approach for this task. Dataset fractal analysis, specifically
the calculation of \emph{intrinsic fractal dimension} (FD) of a dataset,
provides the quantitative means of investigating the non-linearity
and the correlation between the available `features' (i.e., dimensions)
by means of dimensionality of the embedding space \cite{chen_fracanal_1989,moore_crossval_1994}.
Fractal dimension has also been used as an alternative way of characterizing
the discriminative power of each `feature' separately, thus providing
a non-statistical way of ranking them in terms of importance, e.g.
as means of non-parametric feature selection in classification tasks
\cite{Traina_fracfeatsel_2000}. The fractal analysis of datasets
has been used successfully in previous studies \cite{mmhg_mammoejr_2005,mmhg_mammoshape_2007,mmhg_mammotextr_2006}
and it has been proven very valuable as a tool for comparing arbitrary
datasets of extracted features with the qualitative clinical properties
that an experience physician uses to characterize a mammographic image. 

The two most commonly used methods of calculating the fractal dimension
of a dataset are the \emph{pair-count} ($PC$) and the \emph{box-counting}
($BC$) algorithms \cite{fracbookw_1994,moore_crossval_1994,Pierre_fracanalimg_1996,fracbookg_1992}.
In the pair-count algorithm, all Euclidean distances between the samples
of the dataset are calculated and a closure measure is then used to
cluster the resulting distances space into groups, according to various
ranges $r$, i.e., the maximum allowable distance within samples of
the same group. The $PC$ value is calculated for various sizes of
$r$ and it has been proved that $PC\left(r\right)$ can be approximated
by:

\begin{equation}
PC\left(r\right)=K\cdot r^{D}\label{eq:PC}
\end{equation}
where $K$ is a constant and $D$ is called pair-count exponent. The
$PC\left(r\right)$ plot is then a plot of: $log\left(PC\left(r\right)\right)$
versus $log\left(\nicefrac{1}{r}\right)$, i.e., $D$ is the slope
of the linear part of the $PC\left(r\right)$ plot over a specific
range of distances $r$. The exponent $D$ is called \emph{correlation
fractal dimension} of the dataset, or $D_{2}$. 

The box-counting approach calculates the exponent $D$ in a slightly
different way, in order to accommodate case of large datasets with
size in the order of thousands; however, it essentially calculates
an approximation of that same correlation fractal dimension value,
i.e., $D_{2}$. It is commonly used when the datasets contain large
number of samples, usually in the order of thousands \cite{Traina_fracfeatsel_2000,Abrahao_fracdset_2003}.
In this case, instead of calculating all distances between the samples,
the input space is partitioned into a grid of $n$-dimensional cells
of side equal to $r$. Then, the samples inside each cell are calculated
and the frequency of occurrence $R_{r}$, i.e., the count of samples
in a cell, divided by the total number of samples, is used to approximate
the correlation fractal dimension by:
\begin{equation}
D_{2}=\frac{\partial log\underset{i}{\sum}\left(R_{r}^{i}\right)^{2}}{\partial log\left(\nicefrac{1}{r}\right)}\label{eq:corr-fracdim}
\end{equation}

Ideally, both pair-count algorithm and box-counting algorithm calculate
the same value, i.e., the correlation fractal dimension $D_{2}$ of
the initial dataset, which characterizes the intrinsic (true) dimension
of the input space \cite{Abrahao_fracdset_2003}. In other words,
$D_{2}$ would be the \emph{minimum dimension of the dataset} if only
`perfect' features were allowed, i.e., totally uncorrelated and with
the best discriminative power available within the specific set of
features. 

In this study, fractal feature analysis was applied to both the initial
set of qualitative characteristics, provided by the expert physician,
as well as the constructed datasets of morphological features. In
all cases, the pair-count algorithm employing Euclidean distances
was used, due to the relatively small number of samples available,
as well as the better stability and accuracy for $D_{2}$ against
the box-counting approach \cite{Pierre_fracanalimg_1996}. 

In order to calculate the slope at the linear part of the $PC\left(r\right)$
plot, a parametric sigmoid function was used for fitting between the
sample points of the plot. In the parametric sigmoid function: 

\begin{equation}
y=y_{0}+C_{y}\left(\frac{1}{1+exp\left(-C_{x}\left(x-x_{0}\right)\right)}\right)\label{eq:param-sigmoid}
\end{equation}
the $(x_{0},y_{0})$ identifies the transposition of the axes, while
$C_{x}$ and $C_{y}$ identify the appropriate scaling factors. Specifically,
the value of $C_{x}$ affects the steepness of the central part of
the curve, while $C_{y}$ specifies the $Y$-axis width of the sigmoid
curve. Then, the slope of the linear part around the central curvature
point, i.e. the value of $D_{2}$, is:

\begin{equation}
\frac{\partial^{2}y\left(x_{0}\right)}{\partial x^{2}}=0\Rightarrow D_{2}=\frac{\partial y\left(x_{0}\right)}{\partial x}=\frac{C_{x}\cdot C_{y}}{4}\label{eq:param-sigmoid-slope}
\end{equation}

The fitness of the parametric sigmoid over a range of samples assumes
uniform error weighting over the entire range of data. Thus, if a
large percentage of points lies near the upper bound ($y=y_{max}$)
or lower bound ($y=y_{min}$) of the $Y$-axis range, as in most cases
of $PC(r)$ plots, then the fitness in the central region of the sigmoid,
i.e., where the slope is calculated, can be fairly poor. For this
reason, an additional weighting factor was introduced in the fitness
calculation in this study. Specifically, the Tukey (tapered cosine)
parametric window function \cite{Harris_tukey_1978} was applied over
the $Y$-axis range when calculating the overall fitness error of
the sigmoid. The Tukey window is parametric ($q$-value) in terms
of the exact form around its center, ranging from completely rectangular
($q=0$) to completely triangular or Hanning window ($q=1$) . When
applied over the $Y$-axis range, the rectangular case is equivalent
to calculating the fitness error uniformly over the entire range,
while the triangular case is equivalent to calculating the fitness
error primarily against the central point of the sigmoid curve. In
this study, all fitness calculations employed Tukey windows as error
weighting factors, using parameters $q$ in the range between 0.5
and 1.0 for optimal slope results. The equation for computing the
coefficients $w_{j}$ of a discrete Tukey window of length $N$ ($j=1...N$)
is as follows:
\begin{equation}
w_{j}=\begin{cases}
\begin{array}{ccc}
\frac{1}{2}\left(1+cos\left(\frac{2\pi\left(j-1\right)}{q\left(N-1\right)}-\pi\right)\right) & , & 1\leq j<\frac{q}{2}\left(N-1\right)\\
1 & , & \frac{q}{2}\left(N-1\right)\leq j\leq N-\frac{q}{2}\left(N-1\right)\\
\frac{1}{2}\left(1+cos\left(\frac{2\pi}{q}-\frac{2\pi\left(j-1\right)}{q\left(N-1\right)}-\pi\right)\right) & , & N-\frac{q}{2}\left(N-1\right)<j\leq N
\end{array}\end{cases}\label{eq:tukey}
\end{equation}

\subsection{Independent Component Analysis (ICA) \label{sub:ICA-descr}}

In blind source separation (BSS), ICA has been successfully applied
to fMRI data for many years \cite{hyvarbook,calhoun_stica_2001,jung_introduction_2001,correa_comparison}.
Since the fMRI consists of a mixture of unknown components, corresponding
to different brain sources of activity, the unmixing procedure is
essentially a BSS problem. However, due to the relatively low temporal
and spatial resolution of fMRI data, the non-stationary properties
of the signal due to brain- and machine-state variations, as well
as the unknown number and exact statistical properties of the sources,
the BSS of fMRI data is not a trivial task.

ICA is based on identifying non-Gaussian properties between the sources
and separating them from the mixture, essentially reconstructing the
original signal as a linear combination of identified components,
i.e., similarly to the previously discussed formulation $\bm{Y}=\bm{T}\bm{S}$,
$\bm{T}\in\mathbb{R}^{t\times p}$, $\bm{S}\in\mathbb{R}^{p\times n}$.
In this case, $\bm{S}$ is the matrix of independent components (spatial
maps of brain activity) and $\bm{T}$ is the mixture matrix (corresponding
time courses). In fMRI, the ICA can be performed in the spatial or
temporal dimension of the (vectorized) voxel data, producing either
spatial or temporal components in matrix $\bm{S}$. Several studies
have been conducted in whether spatial or temporal ICA works better
for BSS in fMRI data \cite{calhoun_stica_2001}; however spatial maps,
i.e., retrieving $\bm{S}$ as spatial components, seem to be more
accurate and useful in most clinical applications of fMRI. The two
most common approaches for ICA are the Infomax \cite{bell_infomax_1995}
and fastICA \cite{Hyvarinen97afast,Hyvarinen99afast,hyvarbook} algorithms.

Although ICA has been widely studied and employed in fMRI, recent
works have identified the relevant advantages of analyzing brain activity
under the sparsity, instead of statistical independency, of the underlying
mixture of individual components. Additionally, the BSS problem itself
has been identified by few researchers as equivalent to Dictionary
Learning (DL) \cite{abolghasemi_blind_2012,abolghasemi_fast_2013,kangjoo_lee_data-driven_2011,Lee_ddfmri_2010,Eusipco2014},
which is already used in various applications.

Since ICA does not include any sparsity constraints (like in DL),
while at the same time it assumes specific statistical properties
for the underlying signal sources (at most one Gaussian distribution,
minimal noise artifacts). Hence, ICA unmixing of fMRI data that do
not fully satisfy these constraints will construct factorizations
that include the maximum allowable number of components for the reconstruction
of the original (mixed) data with the minimum error. In other words,
as described in section \ref{sub:decimation-dimension}, when the
fMRI data include non-trivial mixtures of sources (as in the case
of the simulated dataset, see section \ref{sub:Simulated-fMRI-datasets}),
ICA will construct a factorization model $\bm{Y}\simeq\bm{T}\bm{S}$,
$\bm{T}\in\mathbb{R}^{t\times p}$, $\bm{S}\in\mathbb{R}^{p\times n}$,
with $p=p_{max}$ and non-zero reconstruction error. Similar problems
emerge when using sparsity-aware approaches as in DL \cite{Eusipco2014},
since they typically produce factorizations with $p\ll p_{max}$ (here,
$p_{max}$ is the dictionary size), but with larger reconstruction
errors, as expected.

In this study, ICA is used as one of the most popular approaches in
BSS problems like the fMRI unmixing task. In the simulated fMRI datasets
(see section \ref{sub:Simulated-fMRI-datasets}), ICA provides an
exact estimation of the intrinsic dimensionality of the signal, which
is expected to be lower than the pre-defined sources used in the mixture.
In the real fMRI datasets (see sections \ref{sub:ds101-descr} and
\ref{sub:ds105-descr}), ICA provides approximate factorization models
and a quantitative way to track the signal reconstruction error as
the number of used components changes. In both cases, the factorization
models provided by ICA are used as a verification tool for validating
the quality of the estimated fractal dimension of each dataset. Although
the exact numbers between $p_{max}$ and the fractal dimension calculated
differ due to their inherently different meaning, tracking their changes
in parallel and comparing results is used here as a valuable tool
for dimensionality analysis of the fMRI datasets.

\section{Datasets}

The investigation of fMRI space complexity and intrinsic dimensionality
was conducted with two separate types datasets, namely one of simulated
fMRI data and two of real fMRI data from carefully designed experiments.
The simulated data were introduced as the means to verify the recoverability
of the intrinsic dimension when the real signal sources are known
and well-defined, while the real data were used as guidelines for
estimating the true brain activity in two typical cognitive tasks
(visual recognition task and visuo-motor task).

\subsection{Simulated fMRI datasets \label{sub:Simulated-fMRI-datasets}}

In this study, an adapted version of the real-valued fMRI data generator
code from the MSLP-Lab \cite{mlsp-lab-url} toolbox was used for creating
artificial fMRI data as a mixture of eight main sources \cite{correa_comparison}.
Using the basic knowledge of the underlying statistical characteristics
of the underlying sources, the components include three highly super-Gaussian
sources (S1, S2, S5), a Gaussian source (S4) and a sub-Gaussian source
(S3), plus two more super-Gaussian sources (S6, S8) and a sub-Gaussian
source (S7). The time course for each component defines the temporal
characteristics of the corresponding source, namely one task-related
(S1), two transiently task-related (S2, S6) and several artifact types
(S3, S4, S5, S7, S8), including respiration, cardiac pulsation, scanner
drift, background noise, etc. These sources can be considered as spatial
maps that are activated according to their time course and mixed linearly
to produce the final (simulated) fMRI data.

Although in typical fMRI experiments there is only one sensory `input'
(stimulation), here the full set of eight sources (S1...S8) was considered
throughout the evaluation. Specifically, the simulated fMRI data included
eight spatial maps of size 60x60 voxels (2-D `slices') and a 100-point
time course, with statistical properties as described above. Each
spatial map was linearized by row-concatenation into a (row) vector
of 3600 voxels, registered along its time course (column) vector of
100 points. Finally, these eight 100x3600 matrices of spatio-temporal
maps were mixed linearly to produce the final eight-source mixing
of simulated fMRI data into one matrix of that same size. Hence, in
terms of the problem formulation presented in section \ref{sub:intro-brain-activity},
the final matrix of (simulated) fMRI data is registered as $\bm{Y}\in\mathbb{R}^{t\times n}$,
where $t=100$ time points and $n=60^{2}=3600$ voxels. Since the
final data matrices are always linearized in a similar way before
applying any unmixing algorithm, using 2-D `slices' of (simulated)
voxels instead of full 3-D (real) brain scans in each time point affects
only the volume of the data and not the task itself.

\subsection{Real fMRI datasets}

\subsubsection{ds101 -- The `Simon' task \label{sub:ds101-descr}}

The `NYU Simon Task' dataset \cite{ds101_simon_2011} comprises of
data collected from 21 healthy adults while they performed a rapid
event-related Simon task. On each trial, the inter-trial interval
(ITI) was 2.5 seconds, with null events for jitter), a red or green
box appeared on the right or left side of the screen. Participants
used their left index finger to respond to the presentation of a green
box, and their right index finger to respond to the presentation of
a red box. In congruent trials the green box appeared on the left
or the red box on the right, while in more demanding incongruent trials
the green box appeared on the right and the red on the left. Subjects
performed two blocks, each containing 48 congruent and 48 incongruent
trials, presented in a pre-determined order (as per OptSeq), interspersed
with 24 null trials (fixation only).

Functional imaging data were acquired using a research dedicated Siemens
Allegra 3.0 T scanner, with a standard Siemens head coil, located
at the NYU Center for Brain Imaging. The data obtained were 151 contiguous
echo planar imaging (EPI) whole-brain functional volumes (TR=2000
ms; TE=30 ms; flip angle=80, 40 slices, matrix=64x64; FOV=192 mm;
acquisition voxel size=3x3x4 mm\textsuperscript{3}) during each of
the two Simon task blocks. A high-resolution T1-weighted anatomical
image was also acquired using a magnetization prepared gradient echo
sequence (MPRAGE, TR=2500 ms; TE= 3.93 ms; TI=900 ms; flip angle=8;
176 slices, FOV=256 mm).

The `ds101' dataset is hosted by Openfmri.org for public access in
raw NiFTI format \cite{nifti_info}, including voxel masks and brain
map template, but without any pre-processing (head movement, sensor
drift, etc). 

In this study, the data from nine (out of 21) subjects were used,
including two runs each, for a total of 18 datasets of fMRI scans.
Each dataset was masked for exclusion of non-brain areas and thresholded
for exclusion of brain areas with near-zero activity. The resulting
number of voxels ranged roughly between 28K and 39K, while the number
of snapshots was fixed to 151 time points. In terms of the formulation
of section \ref{sub:intro-brain-activity}, each fMRI data matrix
is $\bm{Y}\in\mathbb{R}^{t\times n}$ with $t=151$ time points and
$27631\leq n\leq38735$ `non-zero' voxels.

Three variants of each dataset were used, regarding the smoothing
pre-filtering. Specifically, according to standard fMRI acquisition
practice, a Gaussian smoothing kernel was applied to the original
3-D voxel space, in order to suppress noise artifacts and enhance
the spatial continuity of the voxel data. With respect to their \emph{Full
Width at Half Maximum} (FWHM) \cite{Chung_fwhm_2012,fwhmbook_2014},
or $2\sqrt{2\cdot ln2}\cdot\sigma\simeq2.35482\cdot\sigma$ for Gaussian
kernels, two different spatial sizes were used: 4 mm\textsuperscript{3}
and 8 mm\textsuperscript{3}. In practice, since the voxel resolution
in this dataset is 3x3x4 mm\textsuperscript{3}, the smaller kernel
performs (softer) averaging on 1-1.33 neighboring voxels, while the
larger kernel performs (more aggressive) averaging on 2-2.67 neighboring
voxels. These two `smoothed' versions, plus the original non-smoothed
version, are the three variants of each dataset, used throughout the
experiments (see section \ref{sec:Experim-Results} for details).

\subsubsection{ds105 -- Visual object recognition task \label{sub:ds105-descr}}

The `Visual Object Recognition Task' dataset \cite{ds105_visual_2011,Hanson_ds105_2004,Haxby_ds105_2001,OToole_ds105_2005}
comprises of data collected from six healthy adults while they performed
a visual recognition task. Neural responses, as reflected in hemodynamic
changes, were measured in six subjects (five female and one male)
with gradient echo echoplanar imaging (EPI) on a GE 3T scanner (General
Electric, Milwaukee, WI) (repetition time (TR) = 2500 ms, 40 3.5-mm-thick
sagittal images, field of view (FOV) = 24 cm, echo time (TE) = 30
ms, flip angle = 90), while they performed a one-back repetition detection
task. High-resolution T1-weighted spoiled gradient recall (SPGR) images
were obtained for each subject to provide detailed anatomy (124 1.2-mm-thick
sagittal images, FOV = 24 cm). 

Stimuli were gray-scale images of faces, houses, cats, bottles, scissors,
shoes, chairs, and nonsense patterns. The categories were chosen so
that all stimuli from a given category would have the same base level
name. The specific categories were selected to allow comparison with
our previous studies (faces, houses, chairs, animals, and tools) or
ongoing studies (shoes and bottles). Control nonsense patterns were
phase-scrambled images of the intact objects. Twelve time series were
obtained in each subject. Each time series began and ended with 12
s of rest and contained eight stimulus blocks of 24-s duration, one
for each category, separated by 12-s intervals of rest. Stimuli were
presented for 500 ms with an interstimulus interval of 1500 ms. Repetitions
of meaningful stimuli were pictures of the same face or object photographed
from different angles. Stimuli for each meaningful category were four
images each of 12 different exemplars.

The `ds105' dataset is hosted by Openfmri.org for public access in
raw NiFTI format \cite{nifti_info}, including voxel masks and brain
map template, but without any pre-processing (head movement, sensor
drift, etc).

In this study, the data from six (all) subjects were used, including
three (out of 12) runs each, for a total of 18 datasets of fMRI scans.
Each dataset was masked for exclusion of non-brain areas and thresholded
for exclusion of brain areas with near-zero activity. The resulting
number of voxels ranged roughly between 22K and 47K, while the number
of snapshots was fixed to 121 time points. In terms of the formulation
of section \ref{sub:intro-brain-activity}, each fMRI data matrix
is $\bm{Y}\in\mathbb{R}^{t\times n}$ with $t=121$ time points and
$22387\leq n<47192$ `non-zero' voxels.

Three variants of each dataset were used, regarding the smoothing
pre-filtering. Specifically, according to standard fMRI acquisition
practice, a Gaussian smoothing kernel was applied to the original
3-D voxel space, in order to suppress noise artifacts and enhance
the spatial continuity of the voxel data. With respect to their \emph{Full
Width at Half Maximum} (FWHM) \cite{Chung_fwhm_2012,fwhmbook_2014},
or $2\sqrt{2\cdot ln2}\cdot\sigma\simeq2.35482\cdot\sigma$ for Gaussian
kernels, two different spatial sizes were used: 4 mm\textsuperscript{3}
and 8 mm\textsuperscript{3}. In practice, since the voxel resolution
in this dataset is 3x3x3.5 mm\textsuperscript{3}, the smaller kernel
performs (softer) averaging on 1.14-1.33 neighboring voxels, while
the larger kernel performs (more aggressive) averaging on 2.29-2.67
neighboring voxels. These two `smoothed' versions, plus the original
non-smoothed version, are the three variants of each dataset, used
throughout the experiments (see section \ref{sec:Experim-Results}
for details).

\section{Experiments and Results \label{sec:Experim-Results}}

Two separate sets of experiments were conducted in this study, one
for BSS unmixing via ICA and one for dataset fractal dimension estimation.
Both sets included all three fMRI datasets, namely one of simulated
fMRI data and two of real fMRI data experiments (see sections \ref{sub:Simulated-fMRI-datasets},
\ref{sub:ds101-descr}, \ref{sub:ds105-descr}).

\subsection{ICA analysis of the datasets \label{sub:experim-ICA-analysis}}

The ICA experiments that were conducted with the simulated fMRI data
included two distinct realizations of the dataset, generated by the
same procedure and the same specifications as described in section
\ref{sub:Simulated-fMRI-datasets}. Since the data generation includes
several noise components, the two realizations were used as an additional
verification check to validate that slightly different mixtures of
(artificial) fMRI data do not produce significant differences in the
ICA error-versus-components plots and estimated dataset fractal dimension.

\begin{figure*}
\protect\caption{\label{fig:fastICA-timecourses-example} Ideal (blue) and ICA-recovered
(red) time courses of the eight sources in the simulated fMRI dataset.
Parameter \emph{r} is the correlation coefficient between the original
(ideal) and the recovered time course, \emph{p} is the corresponding
p-value and \emph{rmse} is the matching error. The first component
(upper-left corner) corresponds to the pre-defined external stimuli.}

\centering\includegraphics[width=11cm]{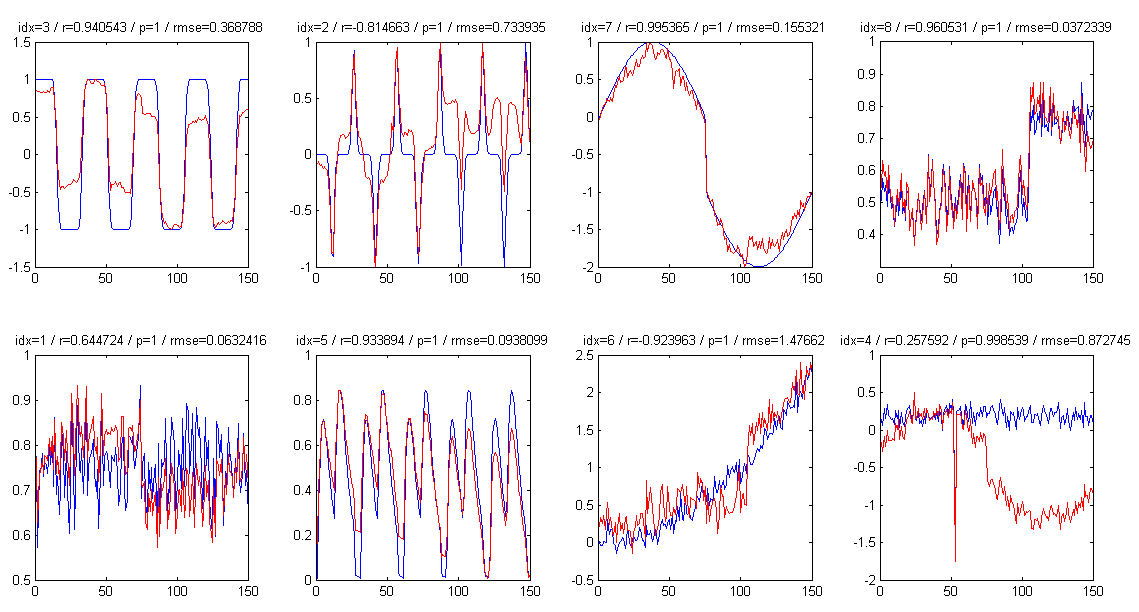}
\end{figure*}

\begin{figure*}
\protect\caption{\label{fig:fastICA-activmaps-example} ICA-recovered activation maps
of the eight sources in the simulated fMRI dataset, spatially reshaped
into proper 2-D brain `slices'. The lower-left box corresponds to
the activation areas for the pre-defined external stimuli. The lower-right
box illustrates the complete reconstructed fMRI mixture at time point
$t=150$.}

\centering\includegraphics[width=9cm]{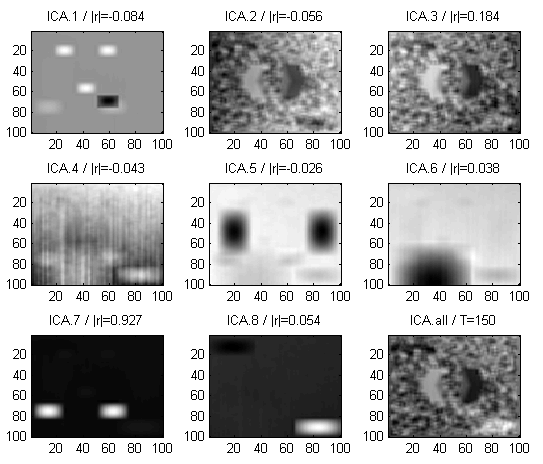}
\end{figure*}

Figure \ref{fig:fastICA-timecourses-example} presents the time courses
of the ICA factorization (matrix $\bm{T}$), with the blue curves
representing each of the eight ideal (true) sources and the red curves
representing the corresponding ICA-recovered sources. Figure \ref{fig:fastICA-activmaps-example}
illustrates the corresponding activation maps (matrix $\bm{S}$) recovered
by ICA, spatially reshaped into proper 2-D brain `slices', where the
reconstruction errors are visible as artifacts (`ghost' artifacts).

\begin{figure*}
\protect\caption{\label{fig:fastICA-simdataset1-ICvsError} Reconstruction error versus
number of used components. ICA detects exactly eight components, i.e.,
the number of true signal sources in the original mixture, and the
final reconstruction error is practically zero.}

\centering\includegraphics[width=10cm]{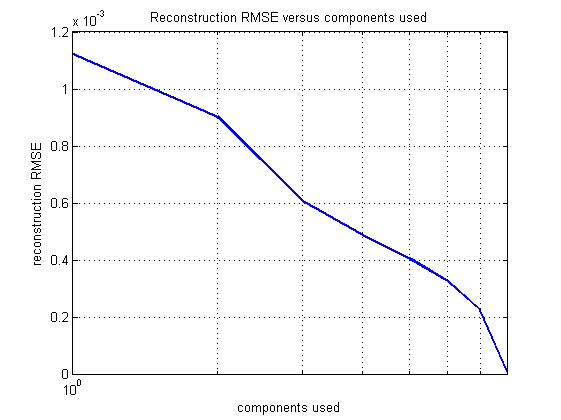}
\end{figure*}

Figure \ref{fig:fastICA-simdataset1-ICvsError} presents the plot
of reconstruction error (RMSE) versus the number of ICA components
used. Specifically, after the ICA unmixing model is complete, the
ICA components are used one by one in rank-1 reconstructions of the
original data and the corresponding errors are used for sorting the
components in ascending order (smallest RMSE first). Subsequently,
a set of components starts from the first one (top of the list) and
increased by adding the next one in each step, while estimating and
registering the corresponding reconstruction error. The plot illustrates
the total reconstruction error decreasing almost linearly as the number
of used components increases, as expected. It should be noted that
for `perfect' ICA factorizations, as in the case of simulated fMRI
data, the number of components recovered by ICA is exactly the same
as the number of signal sources (true) used in the mixture that created
these data (see section \ref{sub:Simulated-fMRI-datasets}).

\begin{figure*}
\protect\caption{\label{fig:3Dmesh-brainslice} Real example of a 2-D brain `slice'
for a single time point, as it is registered in the `ds101' dataset
(unprocessed, no background-exclusion masking).}

\centering\includegraphics[width=11cm]{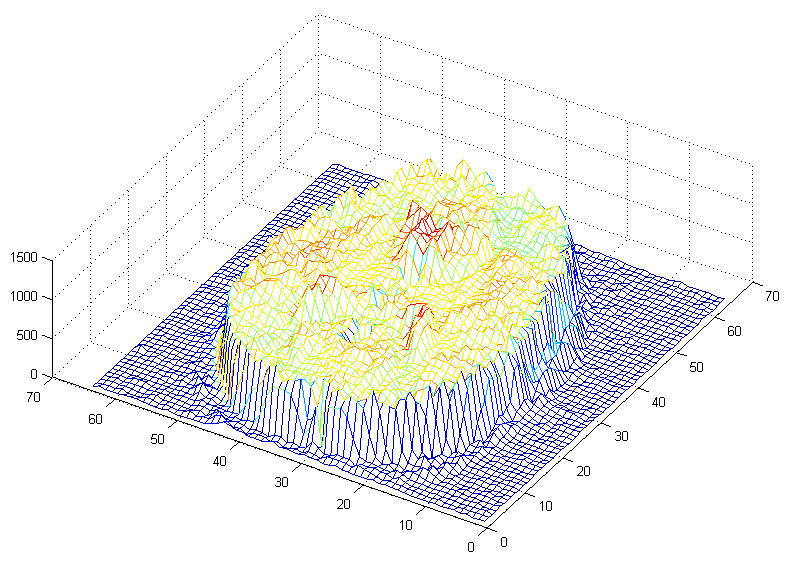}
\end{figure*}

The ICA experiments that were conducted with the real fMRI data included
two distinct datasets, `ds101' and `ds105', as described in sections
\ref{sub:ds101-descr} and \ref{sub:ds105-descr}, respectively. Instead
of a single 2-D brain `slice' as in the case of the simulated fMRI
data, here the datasets employ full 4-D fMRI data, i.e., 3-D voxel
grid of the brain volume evolving in 1-D time course. Figure \ref{fig:3Dmesh-brainslice}
illustrates a real example of a 2-D brain `slice' for a single time
point, as it is registered in the `ds101' dataset; the data are in
raw unprocessed mode and no background-exclusion masking. Signal-independent
noise is evident in the flat/blue areas, i.e., are outside the brain
volume.

\begin{figure*}
\protect\caption{\label{fig:GIFT-ica-example} Sample result from the GIFT toolbox
for Matlab \cite{correa_comparison}, illustrating the ICA-recovered
time course (red plot) and the corresponding 2-D `flattened' activation
map of the actual response of the human brain in a visuo-motor task
very similar to the experimental protocol employed in the `ds101'
dataset.}

\centering\includegraphics[width=11cm]{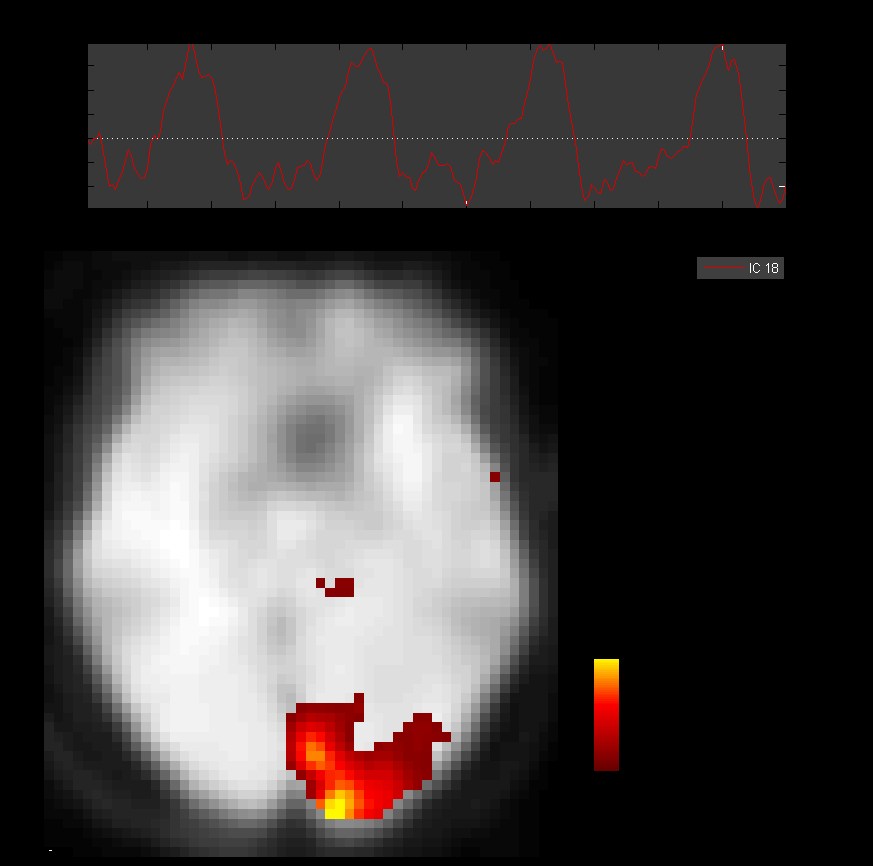}
\end{figure*}

Using the GIFT toolbox for Matlab \cite{correa_comparison}, Figure
\ref{fig:GIFT-ica-example} illustrates the ICA-recovered time course
(red plot) and the corresponding 2-D `flattened' activation map that
represents the actual response of the human brain in a visuo-motor
task very similar to the experimental protocol employed in the `ds101'
dataset. Here, the ICA successfully recovered one particular component
very similar to the external stimuli, which ideally is a square-shaped
pulse modulated by the HRF (see section \ref{sub:intro-fmri-data}),
instead of the noisy sinusoid curve.

\begin{figure*}
\protect\caption{\label{fig:fastICA-realcomponents-example} `ds101' (non-smoothed),
10 of the 50 ICA-recovered time courses of components in a sample
run.}

\centering\includegraphics[width=12cm]{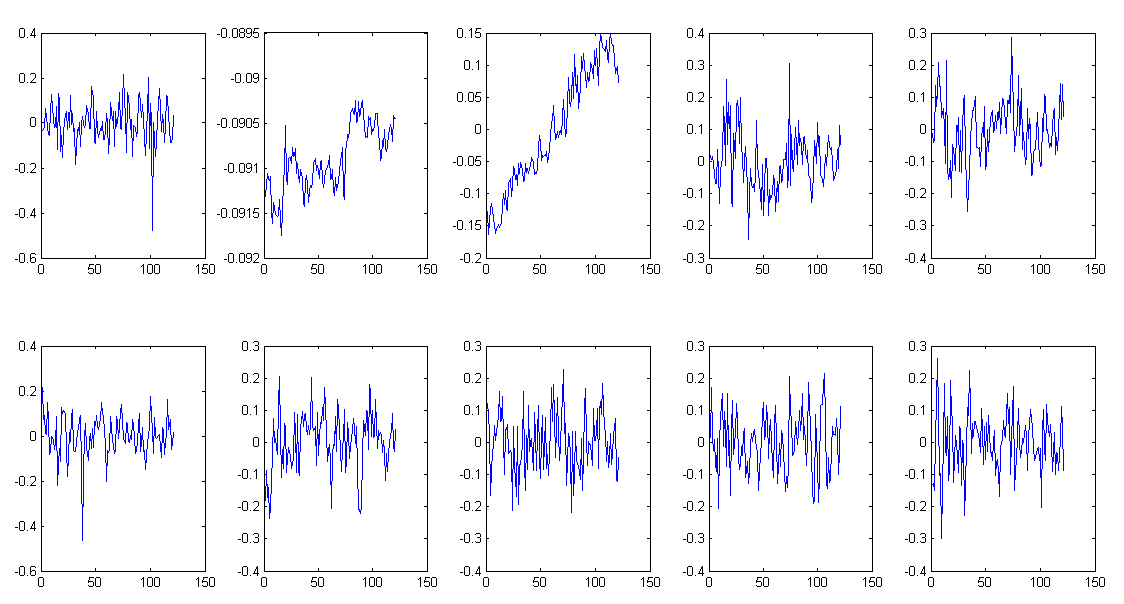}
\end{figure*}

Figure \ref{fig:fastICA-realcomponents-example} illustrates 10 of
the 50 ICA-recovered time courses of components in a sample run with
the `ds101' dataset. Although the ICA converged successfully with
the minimum attainable reconstruction error, the unmixing model failed
to detect one single component that closely matches the ideal time
course of the stimuli. However, it is evident that one component (third
from top-left) matches component no.7 and two components (upper/lower
left) match component no.8 of the simulated fMRI data as illustrated
in Figure \ref{fig:fastICA-timecourses-example} in terms of overall
shape and noise characteristics.

\begin{figure*}
\protect\caption{\label{fig:fastICA-ds101-ICvsError} `ds101' (non-smoothed), ICA reconstruction
error versus number of used components. Red curves represent the RMSE
against the number of used components up to an upper limit of 10,
25, 50 and 100. The final (right-most) point in blue represents the
maximum-size, lowest-RMSE in each case.}

\centering\includegraphics[width=10cm]{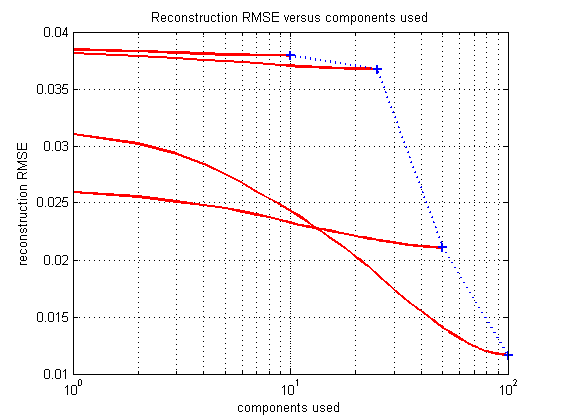}
\end{figure*}

\begin{figure*}
\protect\caption{\label{fig:fastICA-ds105-ICvsError} `ds105' (non-smoothed), ICA reconstruction
error versus number of used components. Red curves represent the RMSE
against the number of used components up to an upper limit of 10,
25, 50 and 100. The final (rightmost) point in blue represents the
maximum-size, lowest-RMSE in each case.}

\centering\includegraphics[width=10cm]{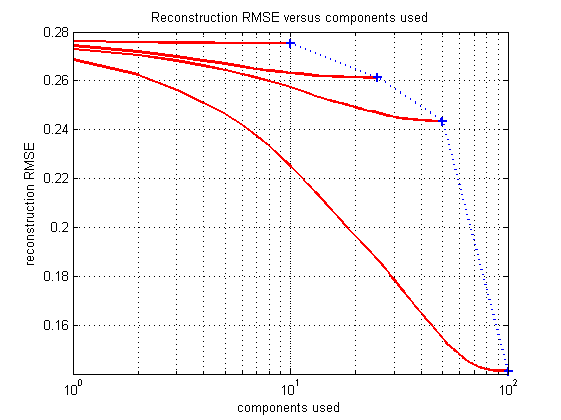}
\end{figure*}

With respect to reconstruction error versus number of used components,
Figure \ref{fig:fastICA-ds101-ICvsError} and Figure \ref{fig:fastICA-ds105-ICvsError}
illustrate how the RMSE changes (drops) as the size of the ICA mixture
becomes larger. Red curves represent the RMSE against the number of
used components up to an upper limit of 10, 25, 50 and 100. The final
(right-most) point in blue represents the maximum-size, lowest-RMSE
in each case. Hence, the general slope of the red curves, as well
as the dotted blue line connecting the end points, illustrate the
robustness of the ICA unmixing process in each of the real fMRI dataset. 

\subsection{Dataset fractal analysis and intrinsic dimensionality \label{sub:experim-FDim-analysis}}

Similarly to the ICA experiments, the dataset fractal analysis that
was conducted with the simulated fMRI data included two distinct realizations
of the dataset, generated by the same procedure and the same specifications
as described in section \ref{sub:Simulated-fMRI-datasets}. Since
the data generation includes several noise components, the two realizations
were used as an additional verification check to validate that slightly
different mixtures of (artificial) fMRI data do not produce significant
differences in the estimation of the fractal dimension of the dataset.

\begin{figure*}
\protect\caption{\label{fig:FDim-simdataset1-loglog} The log-log plot $log\left(PC\left(r\right)\right)$
versus $log\left(\nicefrac{1}{r}\right)$ (blue points) and the best-fit
parametric sigmoid (red curve) that recovers the fractal dimension
(`FDE') of the simulated fMRI dataset. }

\centering\includegraphics[width=9cm]{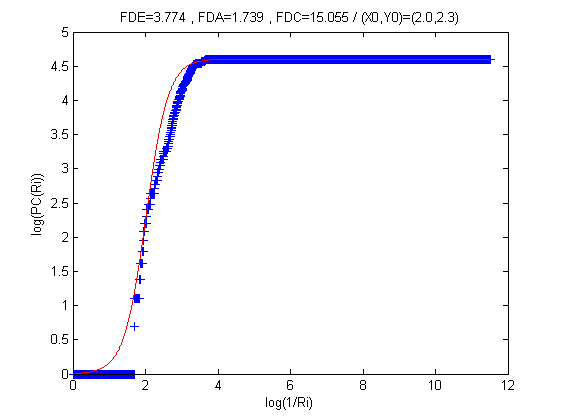}
\end{figure*}

Figure \ref{fig:FDim-simdataset1-loglog} presents the log-log plot
for the box-counting method of estimating the fractal dimension (FD)
in the simulated fMRI dataset, as described in section \ref{sub:dataset-fractal-analysis}.
Specifically, the blue points represent instances of $log\left(PC\left(r\right)\right)$
versus $log\left(\nicefrac{1}{r}\right)$, and the blue curve is the
best-fit parametric sigmoid that is described by Eq.\ref{eq:param-sigmoid}.
The FD is recovered as the slope of the curve in the central point
at $(x_{0},y_{0})$, according to Eq.\ref{eq:param-sigmoid-slope}.

Based on the box-counting approach described in section \ref{sub:dataset-fractal-analysis},
the two realizations of the simulated fMRI data resulted in FD values
of 3.774 and 3.884, using the complete dataset with no decimation
(100 sample vectors). This means that an average mean value of 3.83
can be considered as a reliable estimate of the fractal dimension
of the dataset.

The dataset fractal analysis experiments that were conducted with
the real fMRI data included two distinct datasets, `ds101' and `ds105',
as described in sections \ref{sub:ds101-descr} and \ref{sub:ds105-descr},
respectively. Instead of a single 2-D brain `slice' as in the case
of the simulated fMRI data, here the datasets employ full 4-D fMRI
data, i.e., 3-D voxel grid of the brain volume evolving in 1-D time
course.

\begin{figure*}
\protect\caption{\label{fig:FDsim-ds105sm8-loglog} The log-log plot $log\left(PC\left(r\right)\right)$
versus $log\left(\nicefrac{1}{r}\right)$ (blue points) and the best-fit
parametric sigmoid (red curve) that recovers the fractal dimension
(`FDE') of the `ds105' dataset (smoothed, sm=8mm\textsuperscript{3}).}

\centering\includegraphics[width=9cm]{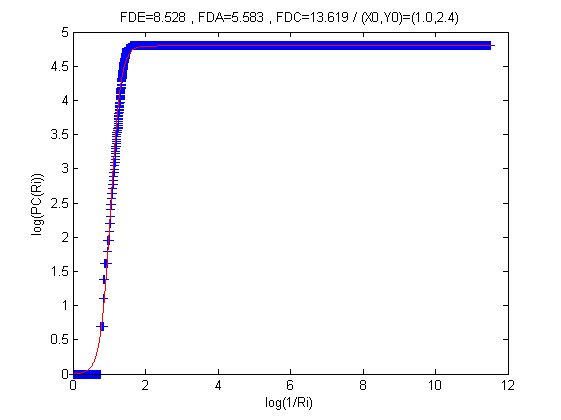}
\end{figure*}

Figure \ref{fig:FDsim-ds105sm8-loglog} presents the log-log plot
for the box-counting method of estimating the fractal dimension (FD)
in the `ds105' dataset, as described in section \ref{sub:ds105-descr};
the corresponding plot for `dsd101' is similar (omitted). Specifically,
the blue points represent instances of $log\left(PC\left(r\right)\right)$
versus $log\left(\nicefrac{1}{r}\right)$, and the blue curve is the
best-fit parametric sigmoid that is described by Eq.\ref{eq:param-sigmoid}.
The FD is recovered as the slope of the curve in the central point
at $(x_{0},y_{0})$, according to Eq.\ref{eq:param-sigmoid-slope}.

Based on the box-counting approach described in section \ref{sub:dataset-fractal-analysis},
the `ds101' and `ds105' datasets were analyzed for multiple subjects
and various choices of smoothing kernel size (see sections \ref{sub:ds101-descr}
and \ref{sub:ds105-descr}, respectively, for details). Table \ref{tab:FDim-ds101}
presents the FD estimations for the `ds101' dataset; Table \ref{tab:FDim-ds105}
presents the FD estimations for the `ds105' dataset. The results include
the mean values and the corresponding confidence range at the significance
level $a=0.05$, as well as the standard deviations. Plain values
correspond to trimmed sets excluding the smallest and largest value,
while values in parentheses correspond to the non-trimmed (complete)
sets. In both tables, each cell corresponds to FD estimation in 18
instances (9x2 for `ds101' and 6x3 for `ds105'). 

\begin{table}
\protect\caption{\label{tab:FDim-ds101} FD estimation in the `ds101' dataset for various
smoothing kernel (sm) sizes. Plain values correspond to trimmed sets
excluding the smallest and largest value, while values in parentheses
correspond to the non-trimmed (complete) sets. Each cell corresponds
to FD estimation in 18 instances (9x2). Confidence range for mean
value is at the significance level $a=0.05$.}

\centering

\begin{tabular}{|c|c|c|c|}
\hline 
 & mean $\left(\mu\right)$ & conf.range $\left(\mu\pm\right)$  & stdev $\left(\sigma\right)$ \tabularnewline
\hline 
\hline 
\emph{(no sm) } & 61.07 \emph{(60.43)}  & 11.73 \emph{(12.57)}  & 23.93 \emph{(27.20)}\tabularnewline
\hline 
sm=4mm\textsuperscript{3}  & 31.92 \emph{(31.59)}  & 5.13 \emph{(5.62)}  & 10.48 \emph{(12.17)}\tabularnewline
\hline 
sm=8mm\textsuperscript{3}  & 11.27 \emph{(17.14)}  & 2.15 \emph{(2.49)}  & 4.39 \emph{(5.38)}\tabularnewline
\hline 
\end{tabular}
\end{table}

\begin{table}
\protect\caption{\label{tab:FDim-ds105} FD estimation in the `ds105' dataset for various
smoothing kernel (sm) sizes. Plain values correspond to trimmed sets
excluding the smallest and largest value, while values in parentheses
correspond to the non-trimmed (complete) sets. Each cell corresponds
to FD estimation in 18 instances (6x3). Confidence range for mean
value is at the significance level $a=0.05$.}

\centering

\begin{tabular}{|c|c|c|c|}
\hline 
 & mean $\left(\mu\right)$ & conf.range $\left(\mu\pm\right)$  & stdev $\left(\sigma\right)$ \tabularnewline
\hline 
\hline 
\emph{(no sm) } & 15.38 \emph{(17.38)}  & 2.86 \emph{(5.56) } & 5.83 \emph{(12.04)}\tabularnewline
\hline 
sm=4mm\textsuperscript{3}  & 12.79 \emph{(13.75) } & 2.20 \emph{(3.31)}  & 4.48 \emph{(7.16)}\tabularnewline
\hline 
sm=8mm\textsuperscript{3}  & 10.67 \emph{(11.01) } & 1.59 \emph{(1.89) } & 3.24 \emph{(4.09)}\tabularnewline
\hline 
\end{tabular}
\end{table}

\section{Discussion \label{sec:Discussion}}

The results presented in section \ref{sub:experim-FDim-analysis},
as well as the ICA unmixing models that were presented in section
\ref{sub:experim-ICA-analysis}, verify that there is indeed a limited
number of activated brain areas during standard cognitive processes.
Since these activations are present simultaneously, they provide a
hint of how many tasks are `running' in the human brain \emph{in parallel}
as part of its every day functionality.

In section \ref{sub:experim-ICA-analysis}, the results from experiments
with simulated fMRI data illustrate the basic unmixing problem for
brain sensory data, which is relevant not only to fMRI but other modalities
too, e.g. in EEG. The results show that ICA can indeed address the
unmixing task with moderate to good performance, especially with regard
to the signal sources related to well-defined external stimuli (see
component no.7 in Figure \ref{fig:fastICA-timecourses-example} and
Figure \ref{fig:fastICA-activmaps-example}). Due to the nature of
ICA and its inherent constraints, not all signal sources can be correctly
identified and, hence, the recovered components do not match the original
ones perfectly; however, if the statistical assertions about the signal
sources are satisfied adequately, the total reconstruction error can
be minimized effectively. For the simulated fMRI dataset, the total
RMSE for the ICA mixture, reconstructing the original signal with
all the recovered (eight) components, is practically zero (see Figure
\ref{fig:fastICA-simdataset1-ICvsError}). The most important results
in this case are: (a) the number of ICA components recovered matches
the number of true sources used to construct the original mixture
and (b) one of the recovered components closely matches (highly correlated)
with the well-defined external stimuli (square-shaped time course).
This is extremely important in real fMRI experimental protocols, where
specific stimuli types are to be correlated to specific brain areas
for constructing `global' brain\emph{ atlases}.

The ICA experiments with the real fMRI datasets `ds101' and `ds105',
described in section \ref{sub:experim-ICA-analysis}, illustrate the
true performance of ICA in constructing factorizations for real brain
data. Here, the data volume is much larger than in the case of simulated
data, since the voxel grid is now 3-D instead of a single 2-D `slice',
while at the same time the inherent statistics are much more complex,
as expected. From Figure \ref{fig:GIFT-ica-example} and Figure \ref{fig:fastICA-realcomponents-example}
it is clear that ICA works as expected, providing `dense' (non-sparse)
unmixing models with satisfactory performance; however, it is not
always clear what is the nature of each of the recovered components
and how they can be interpreted, especially when specific signal sources
are in question other that a pre-defined external stimuli (e.g. scanner
drift, electronic noise, head movements, respiration, cardiac pulsation,
etc). 

As described in sections \ref{sub:ICA-descr} and \ref{sub:decimation-dimension},
in the case of real fMRI datasets the ICA factorization is only approximate
(RMSE is never zero) and the minimum reconstruction error is achieved
only when using the maximum allowable number of components - which,
in turn, is ICA-limited by the number of time points available (i.e.,
$t$ in matrix $\bm{Y}\in\mathbb{R}^{t\times n}$). In other words,
a `perfect' unmixing model in real brain data requires the largest
possible number of components to be retrieved. On the other hand,
from Figure \ref{fig:fastICA-ds101-ICvsError} and Figure \ref{fig:fastICA-ds105-ICvsError}
it is clear that the reconstruction error drops sharply when the number
of used components is much lower than this upper limit. For the `ds101'
dataset, this number seems to be somewhere in $25<p<50$ (see Figure
\ref{fig:fastICA-ds101-ICvsError}), while for the `ds105' dataset
it is $p\simeq50$ (see Figure \ref{fig:fastICA-ds105-ICvsError}).
In both cases, the non-smoothed variants of the datasets were used,
hence there is no loss of fine-detail activations and these estimations
can be considered as realistic and consistent.

With regard to the fractal analysis on the simulated fMRI data, results
in section \ref{sub:experim-FDim-analysis} illustrate the robustness
and consistency of this method. Figure \ref{fig:FDim-simdataset1-loglog}
presents the log-log plot used to estimate the FD in this case, i.e.,
the intrinsic dimension of the dataset, which is calculated as 3.83
($\pm$1.45\%). This value is consistent with the results of other
studies using the same dataset with sparsity-aware realizations of
factorization models \cite{Eusipco2014}, where the estimated sparsity
is clearly lower (6 or less) than the number of signal sources used
in the original mixture. Furthermore, Figure \ref{fig:FDim-simdataset1-loglog}
shows the robustness of the method, with the use of a parametric sigmoid
function and Tukey window, even when the log-log plot does not provide
a clear hint for the selection of the linear part from where the slope
should be extracted.

For the real fMRI datasets, Tables \ref{tab:FDim-ds101} and \ref{tab:FDim-ds105}
present the detailed estimations of the FD for smoothed and non-smoothed
variants. Specifically, Table \ref{tab:FDim-ds101} shows the mean
FD values for the `ds101' dataset, including the confidence range
and the standard deviation. It is clear that, even in the non-smoothed
`noisy' variant, the intrinsic dimension of the space spanned by the
voxel data is much lower ($48<D<63$) than the dimension of the embedding
space ($27K<n<39K$). Furthermore, the value of FD becomes smaller,
as expected, when smoothing is applied to the data prior to the fractal
analysis process. This proves that the method is consistent in terms
of following the decreasing `complexity' of the dataset, as well as
the fact that smoothing the fMRI voxel data can enhance the quality
of the most important information content (major brain activity areas),
with a possible loss in fine details and/or low-level activations.
Hence, smoothing options in fMRI should be carefully selected in relation
to the specifications of each task, i.e., sensitivity versus specificity
requirements. 

Similar comments are valid for the `ds105', according to the results
in Table \ref{tab:FDim-ds105}. In the non-smoothed `noisy' variant,
the intrinsic dimension of the space spanned by the voxel data is,
again, much lower ($12<D<19$) than the dimension of the embedding
space ($22K<n<48K$) and it becomes even smaller, as expected, when
smoothing is applied to the data prior to the fractal analysis process.
Figure \ref{fig:FDsim-ds105sm8-loglog} shows the log-log plot used
to calculate the FD value for the smoothed variant (sm=8mm\textsuperscript{3})
of the dataset, where it is clear that the proposed fractal analysis
method provides a very reliable estimation. Furthermore, it shows
a much better fit in the sigmoid curve, which means that the box-counting
method (see section \ref{sub:dataset-fractal-analysis}) becomes more
reliable, as expected, when the fMRI data are smoothed.

As it was mentioned earlier, this study focuses on the estimation
of the level of \emph{parallelism} when the human brain is performing
complex cognitive tasks. In some very abstract sense, this is not
much different than trying to recover the (minimum) number of actual
`cpu cores' required to `run' all the active cognitive tasks that
are registered in the entire 3-D brain volume while performing a typical
fMRI experimental protocol that includes visuo-motor tasks.

It is very interesting to see that the real fMRI dataset `ds101',
which corresponds to a visuo-motor task, produces much higher estimated
FD values than the corresponding FD values for the `ds105', which
is a much simpler visual recognition-only task. This means that, as
expected, in the second task there is a much lower number of distinct
activated brain areas, hence much fewer independent cognitive tasks
involved, when no motor response is required by the experimental protocol.
This does not mean that the total volume of brain activation is smaller
but rather than fewer functionality components (`sources') are present
\emph{in parallel} when visual recognition is concerned, rather than
when a proper motor response is required by the subject. This is inherently
the casual link to `cpu cores', where several processes are enabled
to run simultaneously in a digital computer. ICA reconstruction plots
show that when the human brain is concerned, this number is not defined
as a strict threshold but rather in a continuous range; when a specific
activation level is defined, a corresponding number of `brain cores'
can be evaluated. However, in real fMRI data, this range seems to
be non-linear and such a number can be retrieved at the point beyond
which adding more components has only marginal impact to the modeled
brain signal (see Figures \ref{fig:fastICA-ds101-ICvsError} and \ref{fig:fastICA-ds105-ICvsError}).

In short, it seems that normal brain functionality, such as in typical
visual or visuo-motor tasks, involves only a limited number of independent
processes that run in parallel. Some of them are related to this specific
task, while others correspond to basic low-level functionality, e.g.
respiration. Although it is difficult to correctly identify and explain
all these components in strictly data-driven approaches (especially
in BSS methods like ICA), the investigation of the number of major
components, in combination with non-parametric dimensionality recovery
methods such as dataset fractal analysis, can provide very useful
hints for developing brain-like technologies and algorithms. 

Current research endeavors like the Human Brain Project (HBP) by EU
\cite{HBP-EU-url} and Brain Research through Advancing Innovative
Neurotechnologies (BRAIN) by USA \cite{BRAIN-USA-url}, as well as
new innovative VLSI technologies like `TrueNorth' project by IBM \cite{Merolla_truenorth_2014,Sebastian_truenorth_2014},
require reliable evaluations of brain activity not only in the structural
but also in the functional level. A typical voxel size of 3x3x3.5-5
mm\textsuperscript{3} corresponds to roughly 2.5-4 million neurons
of several thousands of synapse interconnections each, or $\nicefrac{1}{40000}$
to $\nicefrac{1}{25000}$ of the total brain volume, while the currently
state-of-the-art `TrueNorth' chip provides 1 million artificial neurons
with only 256 synapses each. Hence, the level of true parallelism
in human brain is a design aspect of paramount importance in future
projects.

\section{Conclusion \label{sec:Conclusion}}

This study presents a purely data-driven approach to the estimation
of the level of \emph{parallelism} in human brain. Using fMRI as the
main modality, the human brain activity was investigated through ICA
for BSS, as well as dataset fractal analysis for the estimation of
the intrinsic (true) dimensionality of fMRI data. In some very abstract
sense, this is not much different than trying to recover the (minimum)
number of actual `cpu cores' required to `run' all the active cognitive
tasks that are registered in the entire 3-D brain volume while performing
a typical fMRI experimental protocol that includes visual-only or
visuo-motor tasks.

Analysis of the non-smoothed variants of the real fMRI datasets (i.e.,
no information loss) proved that even when performing complex visuo-motor
tasks, the number of independent brain processes are in the order
of 50 to 60, while it becomes much lower when visual recognition tasks
(no motor response) is concerned. This means that, in theory, an artificial
equivalent of a brain-like cognitive structure may not require a massively
parallel architecture at the level of single neurons, but rather a
properly designed set of limited processes that run in parallel on
a much lower scale. Hence, although current state-of-the-art VLSI
technologies still include very limited features and processing power
when compared to the real human brain, the assertion of employing
actual parallelism level of much lower order can provide useful hints
to future projects.

\subsubsection*{Acknowledgments}

{\small{}The author wishes to thank professor Sergios Theodoridis,
Yiannis Kopsinis and Anastassios Fytsilis, colleagues at the Dept.
of Informatics \& Telecommunications, National Kapodistrian Univ.
of Athens (NKUA/UoA), for their collaboration in related projects
and the useful discussions for works-in-progress in fMRI analysis
and sparse modeling.}{\small \par}

\bibliographystyle{plain}
\bibliography{refs}

\end{document}